\def\year{2018}\relax
\documentclass[letterpaper]{article} 
\usepackage{aaai18}  
\usepackage{times}  
\usepackage{helvet}  
\usepackage{courier}  
\usepackage{url}  
\usepackage{graphicx}  
\usepackage{amssymb}
\usepackage{subfig}
\usepackage{multirow}
\begin{document}
%

\title{A Question-Focused Multi-Factor Attention Network for Question Answering}

\author{Souvik Kundu \and Hwee Tou Ng\\
Department of Computer Science\\
National University of Singapore\\
\{souvik, nght\}@comp.nus.edu.sg\\
}

\maketitle
\begin{abstract}

Neural network models recently proposed for question answering (QA) primarily focus on capturing the passage-question relation. However, they have minimal capability to link relevant facts distributed across multiple sentences which is crucial in achieving deeper understanding, such as performing multi-sentence reasoning, co-reference resolution, etc. They also do not explicitly focus on the question and answer type which often plays a critical role in QA. In this paper, we propose a novel end-to-end question-focused multi-factor attention network for answer extraction. Multi-factor attentive encoding using tensor-based transformation aggregates meaningful facts even when they are located in multiple sentences. To implicitly infer the answer type, we also propose a max-attentional question aggregation mechanism to encode a question vector based on the important words in a question. During prediction, we incorporate sequence-level encoding of the first wh-word and its immediately following word as an additional source of question type information. Our proposed model achieves significant improvements over the best prior state-of-the-art results on three large-scale challenging QA datasets, namely NewsQA, TriviaQA, and SearchQA.

\end{abstract}
\section{Introduction}
\label{sec:intro}
In machine comprehension-based (MC) question answering (QA), a machine is expected to provide an answer for a given question by understanding texts. 
In recent years, several MC datasets have been released. 
\citeauthor{MCTestdata} (\citeyear{MCTestdata}) released a multiple-choice question answering dataset.
\citeauthor{HermannKGEKSB15} (\citeyear{HermannKGEKSB15}) created a large cloze-style dataset using CNN and Daily Mail news articles. Several models \cite{HermannKGEKSB15,chen2016thorough,KadlecSBK16,Kobayashi2016,attn_over_attn,GAR} based on neural attentional and pointer networks \cite{pointer_net} have been proposed since then. 
\citeauthor{RajpurkarZLL16} (\citeyear{RajpurkarZLL16}) released the SQuAD dataset where the answers are free-form unlike in the previous MC datasets.

Most of the previously released datasets are closed-world, i.e., the questions and answers are formulated given the text passages. As such, the answer spans can often be extracted by simple word and context matching. \citeauthor{newsqa} (\citeyear{newsqa}) attempted to alleviate this issue by proposing the NewsQA dataset where the questions are formed only using the CNN article summaries without accessing the full text. As a result, a significant proportion of questions require reasoning beyond simple word matching. 
Two even more challenging open-world QA datasets, TriviaQA \cite{triviaqa} and SearchQA \cite{searchqa}, have recently been released. 
TriviaQA consists of question-answer pairs authored by trivia enthusiasts and independently gathered evidence documents from Wikipedia as well as Bing Web search. 
In SearchQA, the question-answer pairs are crawled from the Jeopardy archive and are augmented with text snippets retrieved from Google search.

Recently, many neural models have been proposed \cite{mpcm_squad,memen,allenai_squad,smu_squad,fastqa_squad,salesforce_squad,cmu_squad}, which mostly focus on passage-question interaction to capture the context similarity for extracting a text span as the answer. 
However, most of the models do not focus on synthesizing evidence from multiple sentences and fail to perform well on challenging open-world QA tasks such as NewsQA and TriviaQA. 
Moreover, none of the models explicitly focus on question/answer type information for predicting the answer. In practice, fine-grained understanding of question/answer type plays an important role in QA. 

In this work, we propose \underline{a}n end-to-end question-focused \underline{m}ulti-factor \underline{a}ttention \underline{n}etwork for \underline{d}ocument-based question \underline{a}nswering (AMANDA), which learns to aggregate evidence distributed across multiple sentences and identifies the important question words to help extract the answer. 
Intuitively, AMANDA extracts the answer not only by synthesizing relevant facts from the passage but also by implicitly determining the suitable answer type during prediction.
The key contributions of this paper are:
\begin{itemize}
\item We propose a multi-factor attentive encoding approach based on tensor transformation to synthesize meaningful evidence across multiple sentences. It is particularly effective when answering a question requires deeper understanding such as multi-sentence reasoning, co-reference resolution, etc.
\item To subsume fine-grained answer type information, we propose a max-attentional question aggregation mechanism which learns to identify the meaningful portions of a question. We also incorporate sequence-level representations of the first wh-word and its immediately following word in a question as an additional source of question type information.


\end{itemize}
\section{Problem Definition}
\label{sec:prob_def}
Given a pair of passage and question, an MC system needs to extract a text span from the passage as the answer. We formulate the answer as two pointers in the passage, which represent the beginning and ending tokens of the answer. Let $\mathcal{P}$ be a passage with tokens $(\mathcal{P}_1, \mathcal{P}_2, \ldots, \mathcal{P}_T)$ and $\mathcal{Q}$ be a question with tokens $(\mathcal{Q}_1,\mathcal{Q}_2, \ldots, \mathcal{Q}_U)$, where $T$ and $U$ are the length of the passage and question respectively. To answer the question, a system needs to determine two pointers in the passage, $b$ and $e$, such that $1 \leq b \leq e \leq T$. The resulting answer tokens will be $(\mathcal{P}_b, \mathcal{P}_{b+1}, \ldots, \mathcal{P}_e)$. 

\section{Network Architecture}
\label{sec:network}
\begin{figure}[t]
\centering
\includegraphics[width=0.45\textwidth]{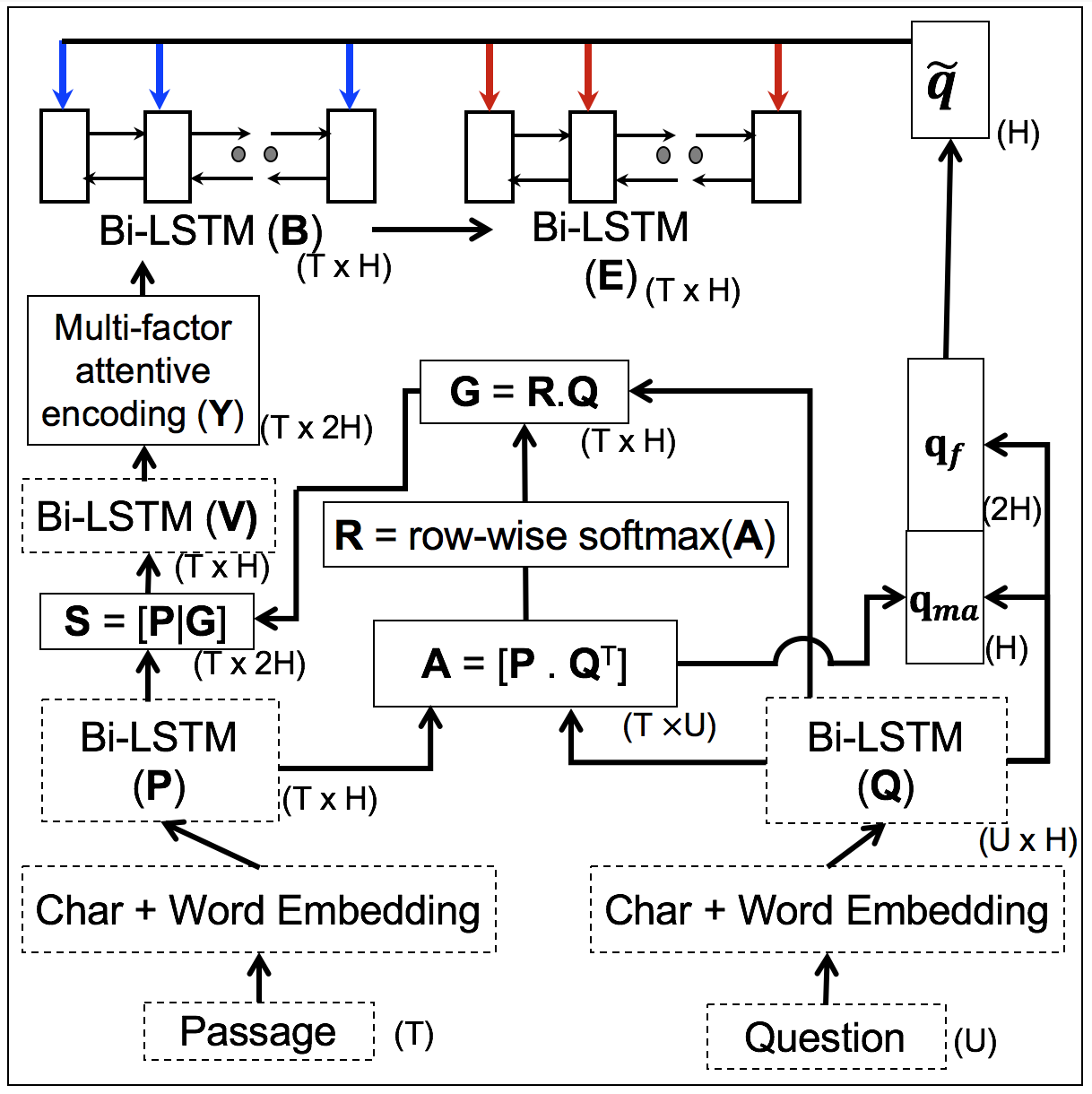}
\caption{Architecture of the proposed model. Hidden unit representations of Bi-LSTMs, \textbf{B} and \textbf{E}, are shown to illustrate the answer pointers. Blue and red arrows represent the start and end answer pointers respectively.}
\label{fig:squad_system_block}
\end{figure}
The architecture of the proposed question-focused multi-factor attention network\footnote{Our code is available at \url{https://github.com/nusnlp/amanda}} 
is given in Figure \ref{fig:squad_system_block}.
\subsection{Word-level Embedding}
Word-level embeddings are formed by two components: pre-trained word embedding vectors from GloVe \cite{pennington2014glove} and convolutional neural network-based (CNN) character embeddings \cite{cnn_char_emb_kim2014}. 
Character embeddings have proven to be very useful for out-of-vocabulary (OOV) words.
We use a character-level CNN followed by max-pooling over an entire word to get the embedding vector for each word. Prior to that, a character-based lookup table is used to generate the embedding for every character and the lookup table weights are learned during training. We concatenate these two embedding vectors for every word to generate word-level embeddings. 
\subsection{Sequence-level Encoding}
We apply sequence-level encoding to incorporate contextual information. Let $\mathbf{e}^p_t$ and $\mathbf{e}^q_t$ be the $t$th embedding vectors of the passage and the question respectively. The embedding vectors are fed to a bi-directional LSTM (BiLSTM) \cite{LSTM}. Considering that the outputs of the BiLSTMs are unfolded across time, we represent the outputs as $\mathbf{P} \in \mathbb{R} ^{T \times H}$ and $\mathbf{Q} \in \mathbb{R} ^{U \times H}$ for passage and question respectively. 
$H$ is the number of hidden units for the BiLSTMs. At every time step, the hidden unit representation of the BiLSTMs is obtained by concatenating the hidden unit representations of the corresponding forward and backward LSTMs. For the passage, at time step $t$, the forward and backward LSTM hidden unit representations can be written as:
\begin{eqnarray}
\label{eq:for_back_lstm}
\overrightarrow{\mathbf{h}}^p_t &=& \overrightarrow{\textnormal{LSTM}}(\overrightarrow{\mathbf{h}}^p_{t-1} ,~ \mathbf{e}^p_t)  \nonumber\\ 
\overleftarrow{\mathbf{h}}^p_t &=& \overleftarrow{\textnormal{LSTM}}(\overleftarrow{\mathbf{h}}^p_{t+1} ,~ \mathbf{e}^p_t) 
\end{eqnarray}
The $t$th row of $\mathbf{P}$ is represented as $\mathbf{p}_t = \overrightarrow{\mathbf{h}}^p_t ~||~ \overleftarrow{\mathbf{h}}^p_t$, 
where $||$ represents the concatenation of two vectors. Similarly, the sequence level encoding for a question is $\mathbf{q}_t = \overrightarrow{\mathbf{h}}^q_t ~||~ \overleftarrow{\mathbf{h}}^q_t$, 
where $\mathbf{q}_t$ is the $t$th row of $\mathbf{Q}$. 
\subsection{Cartesian Similarity-based Attention Layer}
The attention matrix is calculated by taking dot products between all possible combinations of sequence-level encoding vectors for a passage and a question. Note that for calculating the attention matrix, we do not introduce any additional learnable parameters. The attention matrix $\mathbf{A} \in \mathbb{R} ^{T \times U}$ can be expressed as:
\begin{equation}
\mathbf{A} = \mathbf{P} ~\mathbf{Q}^\top  ~~
\end{equation}
Intuitively, $A_{i,j}$ is a measure of the similarity between the sequence-level encoding vectors of the $i$th passage word and the $j$th question word.
\subsection{Question-dependent Passage Encoding}
In this step, we jointly encode the passage and question. We apply a row-wise softmax function on the attention matrix:

\begin{equation}
\mathbf{R} = \textnormal{row-wise softmax} (\mathbf{A})
\end{equation}
If $\mathbf{r}_t \in \mathbb{R} ^{U}$ is the $t$th row of $\mathbf{R} \in \mathbb{R} ^{T \times U}$, then $\sum_{j=1}^{U} r_{t,j} = 1$.
Each row of $\mathbf{R}$ measures how relevant every question word is with respect to a given passage word. Next, an aggregated question vector is computed corresponding to each sequence-level passage word encoding vector. The aggregated question vector $\mathbf{g}_t \in \mathbb{R} ^{H}$ corresponding to the $t$th passage word is computed as $\mathbf{g}_t = \mathbf{r}_t \mathbf{Q}$.  
The aggregated question vectors corresponding to all the passage words can be computed as $\mathbf{G} = \mathbf{R} ~ \mathbf{Q}$,
where $\mathbf{g}_t$ is the $t$th row of $\mathbf{G} \in \mathbb{R} ^{T \times H}$. 

The aggregated question vectors corresponding to the passage words are then concatenated with the sequence-level passage word encoding vectors. If the question-dependent passage encoding is denoted as $\mathbf{S} \in \mathbb{R} ^{T \times 2H}$ and $\mathbf{s}_t$ is the $t$th row of $\mathbf{S}$, then $\mathbf{s}_t = \mathbf{c}_t ~||~ \mathbf{g}_t$, 
where $\mathbf{c}_t$ is the sequence-level encoding vector of the $t$th passage word ($t$th row of $\mathbf{P}$). Then a BiLSTM is applied on $\mathbf{S}$ to obtain $\mathbf{V} \in \mathbb{R} ^{T \times H}$.
\subsection{Multi-factor Attentive Encoding}
Tensor-based neural network approaches have been used in a variety of natural language processing tasks \cite{tensor_neural_net,tensor_discourse}. We propose a multi-factor attentive encoding approach using tensor-based transformation. In practice, recurrent neural networks fail to remember information when the context is long. Our proposed multi-factor attentive encoding approach helps to aggregate meaningful information from a long context with fine-grained inference due to the use of multiple factors while calculating attention. 

Let $\mathbf{v}_{i} \in \mathbb{R} ^{H}$ and $\mathbf{v}_{j} \in \mathbb{R} ^{H}$ represent the question-dependent passage vectors of the $i$th and $j$th word, i.e., the $i$th and $j$th row of $\mathbf{V}$. Tensor-based transformation for multi-factor attention is formulated as follows:
\begin{equation}
\mathbf{f}^m_{i,j} = \mathbf{v}_i ~ \mathbf{W}^{[1:m]}_f ~ \mathbf{v}^\top_j ~~,
\end{equation}
where $\mathbf{W}^{[1:m]}_f \in \mathbb{R} ^{H \times m \times H}$ is a 3-way tensor and $m$ is the number of factors. The output of the tensor product $\mathbf{f}^m_{i,j} \in \mathbb{R} ^{m}$ is a vector where each element $f^m_{i,j,k}$ is a result of the bilinear form defined by each tensor slice $\mathbf{W}^{[k]}_f \in \mathbb{R} ^{H \times H}$:
\begin{equation}
f^m_{i,j,k} = \mathbf{v}_i ~ \mathbf{W}^{[k]}_f ~ \mathbf{v}^\top_j = \sum_{a,b} v_{i,a} W^{[k]}_{f_{a,b}} v_{j,b}
\end{equation}
$\forall i,j \in [1,T]$, the multi-factor attention tensor can be given as $\mathbf{F}^{[1:m]} \in \mathbb{R} ^{m \times T \times T}$. 
For every vector $\mathbf{f}^m_{i,j}$ of $\mathbf{F}^{[1:m]}$, we perform a max pooling operation over all the elements to obtain the resulting attention value:
\begin{equation}
F_{i,j} = \textnormal{max}(\mathbf{f}^m_{i,j}) ~~,
\end{equation}
where $F_{i,j}$ represents the element in the $i$th row and $j$th column of $\mathbf{F} \in \mathbb{R} ^{T \times T}$. Each row of $\mathbf{F}$ measures how relevant every passage word is with respect to a given question-dependent passage encoding of a word. 
We apply a row-wise softmax function on $\mathbf{F}$ to normalize the attention weights, obtaining $\mathbf{\tilde{F}} \in \mathbb{R} ^{T \times T}$.
Next, an aggregated multi-factor attentive encoding vector is computed corresponding to each question-dependent passage word encoding vector. 
The aggregated vectors corresponding to all the passage words, $\mathbf{M} \in \mathbb{R} ^{T \times H}$, can be given as $\mathbf{M} = \mathbf{\tilde{F}} ~ \mathbf{V}$. 
The aggregated multi-factor attentive encoding vectors are concatenated with the question-dependent passage word encoding vectors to obtain $\mathbf{\tilde{M}} \in \mathbb{R} ^{T \times 2H}$. To control the impact of $\mathbf{\tilde{M}}$, we apply a feed-forward neural network-based gating method to obtain $\mathbf{Y} \in  \mathbb{R} ^{T \times 2H}$. If the $t$th row of $\mathbf{\tilde{M}}$ is $\mathbf{\tilde{m}}_t$, then the $t$th row of $\mathbf{Y}$ is:
\begin{equation}
\mathbf{y}_t = \mathbf{\tilde{m}}_t \odot \textnormal{sigmoid}(\mathbf{\tilde{m}}_t \mathbf{W}^g + \mathbf{b}^g) ~~,
\end{equation}
where $\odot$ represents element-wise multiplication.
$\mathbf{W}^g \in \mathbb{R} ^{2H \times 2H}$ and $\mathbf{b}^g \in \mathbb{R} ^{2H}$ are the transformation matrix and bias vector respectively.

We use another pair of stacked BiLSTMs on top of $\mathbf{Y}$ to determine the beginning and ending pointers. Let the hidden unit representations of these two BiLSTMs be $\mathbf{B} \in \mathbb{R} ^{T \times H}$ and $\mathbf{E} \in \mathbb{R} ^{T \times H}$. To incorporate the dependency of the ending pointer on the beginning pointer, the hidden unit representation of $\mathbf{B}$ is used as input to $\mathbf{E}$.
\subsection{Question-focused Attentional Pointing}
Unlike previous approaches, our proposed model does not predict the answer pointers directly from contextual passage encoding or use another decoder for generating the pointers. 
We formulate a question representation based on two parts: 
\begin{itemize}
\item max-attentional question aggregation ($\mathbf{q}_{ma}$) 
\item question type representation ($\mathbf{q}_{f}$)
\end{itemize}
$\mathbf{q}_{ma}$ is formulated by using the attention matrix $\mathbf{A}$ and the sequence-level question encoding $\mathbf{Q}$. We apply a \textnormal{maxcol} operation on $\mathbf{A}$ which forms a row vector whose elements are the maximum of the corresponding columns of $\mathbf{A}$. 
We define $\mathbf{k} \in \mathbb{R} ^{U}$ as the normalized max-attentional weights:
\begin{equation}
\label{eq:maxcolA}
\mathbf{k} = \textnormal{softmax} (\textnormal{maxcol}(\mathbf{A}))
\end{equation}
where \textnormal{softmax} is used for normalization.
The max-attentional question representation $\mathbf{q}_{ma} \in \mathbb{R} ^{H}$ is:
\begin{equation}
\label{eq:q_ma}
\mathbf{q}_{ma} = \mathbf{k} ~ \mathbf{Q}
\end{equation}
Intuitively, $\mathbf{q}_{ma}$ aggregates the most relevant parts of the question with respect to all the words in the passage. 

$\mathbf{q}_{f}$ is the vector concatenation of the representations of the first wh-word and its following word from the sequence-level question encoding $\mathbf{Q}$. The set of wh-words we used is \{{\em what, who, how, when, which, where, why}\}. If $\mathbf{q}_{t_{wh}}$ and $\mathbf{q}_{t_{wh}+1}$ represent the first wh-word and its following word (i.e., the $t_{wh}$th and $(t_{wh}+1)$th rows of $\mathbf{Q}$), $\mathbf{q}_{f} \in \mathbb{R} ^{2H}$ is expressed as:
\begin{equation}
\centering
\label{eq:q_type}
\mathbf{q}_{f} ~=~ \mathbf{q}_{t_{wh}} ~||~ \mathbf{q}_{t_{wh}+1}
\end{equation}

The final question representation $\mathbf{\tilde{q}} \in \mathbb{R} ^{H}$ is expressed as:
\begin{equation}
\label{eq:q_tilde}
\mathbf{\tilde{q}} = \textnormal{tanh} ((\mathbf{q}_{ma} ~||~ \mathbf{q}_{f}) \mathbf{W}_q ~+~ \mathbf{b}_q )
\end{equation}
where $\mathbf{W}_q \in \mathbb{R} ^{3H \times H}$ and $\mathbf{b}_q \in \mathbb{R} ^{H}$ are the weight matrix and bias vector respectively. If no wh-word is present in a question, we use the first two sequence-level question word representations for calculating $\mathbf{\tilde{q}}$.

We measure the similarity between $\mathbf{\tilde{q}}$ and the contextual encoding vectors in $\mathbf{B}$ and $\mathbf{E}$ to determine the beginning and ending answer pointers. Corresponding similarity vectors $\mathbf{s}_b \in \mathbb{R} ^{T}$ and $\mathbf{s}_e \in \mathbb{R} ^{T}$ are computed as:
\begin{equation}
\mathbf{s}_b = \mathbf{\tilde{q}} ~ \mathbf{B}^\top  ~~,~~ \mathbf{s}_e = \mathbf{\tilde{q}} ~ \mathbf{E}^\top
\end{equation}
The probability distributions for the beginning pointer $b$ and the ending pointer $e$ for a given passage $\mathcal{P}$ and a question $\mathcal{Q}$ can be given as:
\begin{eqnarray}
\textnormal{Pr}(b ~|~ \mathcal{P},\mathcal{Q}) &=& \textnormal{softmax}(\mathbf{s}_b) \nonumber \\
\textnormal{Pr}(e ~|~ \mathcal{P},\mathcal{Q}, b) &=& \textnormal{softmax}(\mathbf{s}_e)
\end{eqnarray}
The joint probability distribution for obtaining the answer $a$ is given as:
\begin{equation}
\label{eq:joint_prob}
\textnormal{Pr}(a ~|~ \mathcal{P},\mathcal{Q}) = \textnormal{Pr}(b ~|~ \mathcal{P},\mathcal{Q}) ~ \textnormal{Pr}(e ~|~ \mathcal{P},\mathcal{Q}, b)
\end{equation}

To train our model, we minimize the cross entropy loss:
\begin{equation}
\textnormal{loss} = - \sum \textnormal{log} ~~ \textnormal{Pr}(a ~|~ \mathcal{P},\mathcal{Q})
\end{equation}
summing over all training instances. During prediction, we select the locations in the passage for which the product of $\textnormal{Pr}(b)$ and $\textnormal{Pr}(e)$ is maximum keeping the constraint $1 \leq b \leq e \leq T$.
\begin{table}[t]
\centering
\begin{tabular}{|p{8.0cm}|}
\hline
\textbf{Passage}: ... The family of a Korean-American missionary believed held in North Korea said Tuesday they are working with U.S. officials to get him returned home. Robert Park told relatives before Christmas that he was trying to sneak into the isolated communist state to bring a message of "Christ's love and forgiveness" to North Korean leader Kim ... \\ 
\textbf{Question}: What is the name of the Korean-American missionary? \\ 
\textbf{Reference Answer}: Robert Park \\ \hline
\end{tabular}
\caption{Example of a (passage, question, answer)}
\label{tab:ex2}
\end{table}

\section{Visualization}
\label{sec:visualization}
\begin{figure}[t]
\centering
\includegraphics[width=0.3\textwidth]{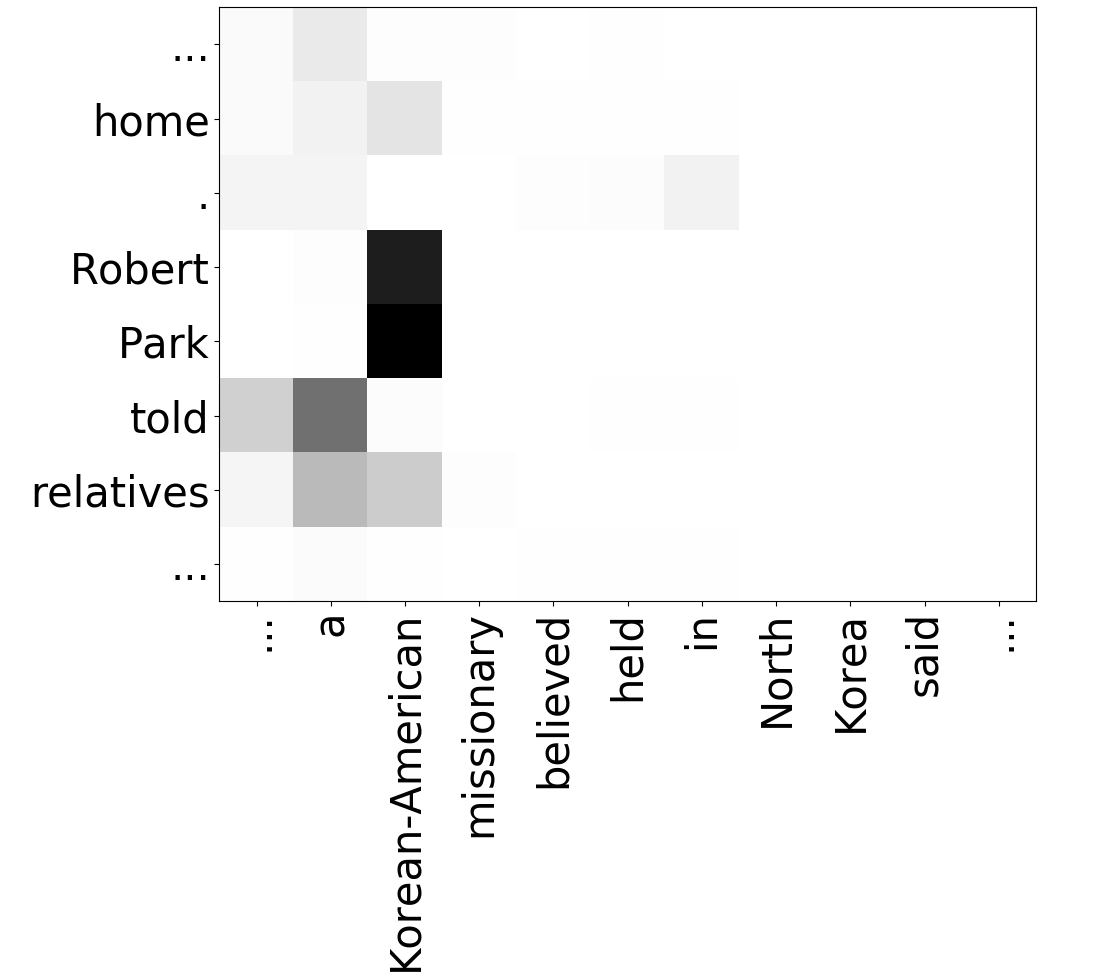}
\caption{Multi-factor attention weights (darker regions signify higher weights).}
\label{fig:ex_mfa}
\end{figure}
\begin{figure}[t]
\centering
\includegraphics[width=5cm,height=3cm]{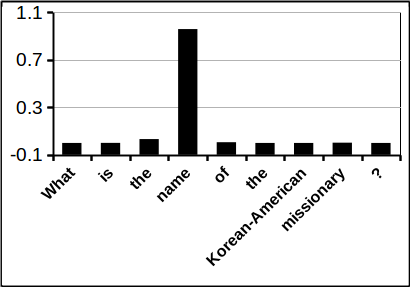}
\caption{Max-attentional weights for question (the origin is set to $-0.1$ for clarity).}
\label{fig:ex_qma}
\end{figure}
To understand how the proposed model works, for the example given in Table \ref{tab:ex2}, we visualize the normalized multi-factor attention weights $\mathbf{\tilde{F}}$ and the attention weights $\mathbf{k}$ which are used for max-attentional question aggregation.

In Figure \ref{fig:ex_mfa}, a small portion of $\mathbf{\tilde{F}}$ has been shown, in which the answer words {\em Robert} and {\em Park} are both assigned higher weights when paired with the context word  {\em Korean-American}. Due to the use of multi-factor attention, the answer segment pays more attention to the important keyword although it is quite far in the context passage and thus effectively infers the correct answer by deeper understanding.
In Figure \ref{fig:ex_qma}, it is clear that the important question word {\em name} is getting a higher weight than the other question words. This helps to infer the fine-grained answer type during prediction, i.e., a person's name in this example.
\section{Experiments}
\label{sec:experiments}
\begin{table}[t]
\centering
\begin{tabular}{l c c c c}
\hline
\multirow{2}{*}{\textbf{Model}} & \multicolumn{2}{c}{\textbf{Dev}} & \multicolumn{2}{c}{\textbf{Test}} \\ \cline{2-5} 
                       & \textbf{EM}          & \textbf{F1}         & \textbf{EM}          & \textbf{F1}          \\ \hline
\multicolumn{3}{l}{\cite{newsqa}} \\
~~Match-LSTM      & 34.4        & 49.6       & 34.9        & 50.0        \\
~~BARB            & 36.1        & 49.6       & 34.1        & 48.2        \\ \hline
\begin{tabular}[l]{@{}l@{}}\cite{two_stage_synnet}\\ ~~BiDAF on NewsQA\end{tabular}      & -        & -       & 37.1        & 52.3     \\ 
\hline
~$^\dagger$Neural BoW  Baseline     & 25.8        & 37.6       & 24.1        & 36.6        \\ 
~$^\dagger$FastQA        & 43.7        & 56.4       & 41.9        & 55.7        \\ 
~$^\dagger$FastQAExt     & 43.7        & 56.1       & 42.8        & 56.1        \\ \hline
\begin{tabular}[l]{@{}l@{}}\cite{reading_twice}\\ ~~R2-BiLSTM\end{tabular}             & -        & -       & 43.7        & 56.7        \\ \hline
\hline
AMANDA     &     \textbf{48.4}    &     \textbf{63.3}     &     \textbf{48.4}    &     \textbf{63.7}   \\ 
\hline
\end{tabular}
\caption{Results on the NewsQA dataset. $^\dagger$ denotes the models of \cite{fastqa_squad}.}
\label{tab:newsqa}
\end{table}
\begin{table*}[t]
\centering
\begin{tabular}{l|c|c|c|c|c|c|c|c|c}
\hline
\multirow{3}{*}{\textbf{Model}} & \multirow{3}{*}{\textbf{Domain}} & \multicolumn{4}{c|}{\textbf{Distant Supervision}} & \multicolumn{4}{c}{\textbf{Verified}} \\ \cline{3-10} 
 &  & \multicolumn{2}{c|}{\textbf{Dev}} & \multicolumn{2}{c|}{\textbf{Test}} & \multicolumn{2}{c|}{\textbf{Dev}} & \multicolumn{2}{c}{\textbf{Test}} \\ \cline{3-10} 
 &  & \textbf{EM} & \textbf{F1} & \textbf{EM} & \textbf{F1} & \textbf{EM} & \textbf{F1} & \textbf{EM} & \textbf{F1} \\ \hline
 \hline
$^\ddagger$Random & \multirow{5}{*}{Wiki} & 12.72 & 22.91 & 12.74 & 22.35 & 14.81 & 23.31 & 15.41 & 25.44 \\  
$^\ddagger$Classifier &  & 23.42 & 27.68 & 22.45 & 26.52 & 24.91 & 29.43 & 27.23 & 31.37 \\ 
$^\ddagger$BiDAF &  & 40.26 & 45.74 & 40.32 & 45.91 & 47.47 & 53.70 & 44.86 & 50.71 \\ 
$^\star$MEMEN &  & 43.16 & 46.90 & - & - & 49.28 & 55.83 & - & - \\ \cline{1-1} \cline{3-10} 
{AMANDA} &  & \textbf{46.95} & \textbf{52.51} & \textbf{46.67} & \textbf{52.22} & \textbf{52.86} & \textbf{58.74} & \textbf{50.51} & \textbf{55.93} \\ \hline
\hline
$^\ddagger$Classifier & \multirow{4}{*}{Web} & 24.64 & 29.08 & 24.00 & 28.38 & 27.38 & 31.91 & 30.17 & 34.67 \\ 
$^\ddagger$BiDAF &  & 41.08 & 47.40 & 40.74 & 47.05 & 51.38 & 55.47 & 49.54 & 55.80 \\ 
$^\star$MEMEN &  & 44.25 & 48.34 & - & - & 53.27 & 57.64 & - & - \\ \cline{1-1} \cline{3-10} 
{AMANDA} &  & \textbf{46.68} & \textbf{53.27} & \textbf{46.58} & \textbf{53.13} & \textbf{60.31} & \textbf{64.90} & \textbf{55.14} & \textbf{62.88} \\ \hline
\end{tabular}
\caption{Results on the TriviaQA dataset. $^\ddagger$\cite{triviaqa}, $^\star$\cite{memen}}
\label{tab:triviaqa}
\end{table*}
\begin{table}[t]
\centering
\begin{tabular}{l|c||c|c}
\hline
\multirow{2}{*}{\textbf{Model}} & \multirow{2}{*}{\textbf{Set}} & \textbf{Unigram} & \textbf{N-gram} \\ 
 &  & \textbf{Accuracy} & \textbf{F1} \\ \hline
\cite{searchqa} & & & \\
\multirow{2}{*}{~~~TF-IDF Max} & Dev & 13.0 & - \\ 
 & Test & 12.7 & - \\ \cline{2-4}
\multirow{2}{*}{~~~ASR} & Dev & 43.9 & 24.2 \\ 
 & Test & 41.3 & 22.8 \\ \hline
 \hline
\multirow{2}{*}{AMANDA} & Dev & \textbf{48.6} & \textbf{57.7} \\ 
 & Test & \textbf{46.8} & \textbf{56.6} \\ \hline
\end{tabular}
\caption{Results on the SearchQA dataset.}
\label{tab:results_searchqa}
\end{table}
We evaluated AMANDA on three challenging QA datasets: NewsQA, TriviaQA, and SearchQA. Using the NewsQA development set as a benchmark, we perform rigorous analysis for better understanding of how our proposed model works.
\subsection{Datasets}
\label{sec:dataset}
The NewsQA dataset \cite{newsqa} consists of around 100K answerable questions 
in total. Similar to \cite{newsqa,fastqa_squad}, we do not consider the unanswerable questions in our experiments. 
NewsQA is more challenging compared to the previously released datasets as a significant proportion of questions requires reasoning beyond simple word- and context-matching. This is due to the fact that the questions in NewsQA were formulated only based on summaries without accessing the main text of the articles. Moreover, NewsQA passages are significantly longer (average length of 616 words) and cover a wider range of topics.

TriviaQA \cite{triviaqa} consists of question-answer pairs authored by trivia enthusiasts and independently gathered evidence documents from Wikipedia and Bing Web search. This makes the task more similar to real-life IR-style QA. In total, the dataset consists of over 650K question-answer-evidence triples. Due to the high redundancy in Web search results (around 6 documents per question), each question-answer-evidence triple is treated as a separate sample and evaluation is performed at document level. However, in Wikipedia, questions are not repeated (each question has 1.8 evidence documents) and evaluation is performed over questions. In addition to distant supervision, TriviaQA also has a verified human-annotated question-evidence collection. Compared to previous datasets, TriviaQA has more complex compositional questions which require greater multi-sentence reasoning.

SearchQA \cite{searchqa} is also constructed to more closely reflect IR-style QA. They first collected existing question-answer pairs from a Jeopardy archive and augmented them with text snippets retrieved by Google.
One difference with TriviaQA is that the evidence passages in SearchQA are Google snippets instead of Wikipedia or Web search documents. This makes reasoning more challenging as the snippets are often very noisy. SearchQA consists of 140,461 question-answer pairs, where each pair has 49.6 snippets on average and each snippet has 37.3 tokens on average. 


\subsection{Experimental Settings}
\label{sec:exp_set}
We tokenize the corpora with NLTK\footnote{http://www.nltk.org/}. We use the 300-dimension pre-trained word vectors from GloVe \cite{pennington2014glove} and we do not update them during training. The out-of-vocabulary words are initialized with zero vectors. We use 50-dimension character-level embedding vectors. 
The number of hidden units in all the LSTMs is 150. We use dropout \cite{dropout} with probability 0.3 for every learnable layer. 
For multi-factor attentive encoding, we choose 4 factors ($m$) based on our experimental findings (refer to Table \ref{tab:variation_num_factors}).
During training, the minibatch size is fixed at 60. 
We use the Adam optimizer \cite{adam} with learning rate of 0.001 and clipnorm of 5. 
During testing, we enforce the constraint that the ending pointer will always be equal to or greater than the beginning pointer. We use exact match (EM) and F1 scores as the evaluation metrics.  
\subsection{Results}
\label{sec:results}
Table \ref{tab:newsqa} shows that AMANDA outperforms all the state-of-the-art models by a significant margin on the NewsQA dataset. 
Table \ref{tab:triviaqa} shows the results on the TriviaQA dataset. In Table \ref{tab:triviaqa}, the model named {Classifier} based on feature engineering was proposed by \citeauthor{triviaqa} (\citeyear{triviaqa}). They also reported the performance of BiDAF \cite{allenai_squad}. A memory network-based approach, {MEMEN}, was recently proposed by \cite{memen}. 
Note that in the Wikipedia domain, we choose the answer which provides the highest maximum joint probability (according to Eq. (\ref{eq:joint_prob})) for any document.
Table \ref{tab:triviaqa} shows that AMANDA achieves state-of-the-art results in both Wikipedia and Web domain on distantly supervised and verified data. 

Results on the SearchQA dataset are shown in Table \ref{tab:results_searchqa}. In addition to a TF-IDF approach, \citeauthor{searchqa} (\citeyear{searchqa}) modified and reported the performance of attention sum reader (ASR) which was originally proposed by \citeauthor{KadlecSBK16} (\citeyear{KadlecSBK16}). 
We consider a maximum of 150 words surrounding the answer from the concatenated ranked list of snippets as a passage to more quickly train the model and to reduce the amount of noisy information.
During prediction, we choose the first 200 words (about 5 snippets) from the concatenated ranked list of snippets as an evidence passage. These are chosen based on performance on the development set.
Based on question patterns, question types are always represented by the first two sequence-level representations of question words. 
To make the results comparable, we also report accuracy for single-word-answer (unigram) questions and F1 score for multi-word-answer (n-gram) questions. 
AMANDA outperforms both systems, especially for multi-word-answer questions by a huge margin. This indicates that AMANDA can learn to make inference reasonably well even if the evidence passages are noisy.

\subsection{Effectiveness of the Model Components}
\label{sec:ablation}
\begin{table}[t]
\centering
\begin{tabular}{l|c|c}
\hline
\textbf{Model} & \textbf{EM} & \textbf{F1} \\ \hline
minus multi factor attn. & 46.4 & 61.2 \\ 
minus $\mathbf{q}_{ma}$ and $\mathbf{q}_{f}$ & 46.2 & 60.5 \\
minus $\mathbf{q}_{ma}$ & 46.6 & 61.3 \\ 
minus $\mathbf{q}_{f}$ & 46.8 & 61.8 \\ \hline
\hline
{AMANDA} & \textbf{48.4} & \textbf{63.3} \\ \hline
\end{tabular}
\caption{Ablation of proposed components on the NewsQA development set.}
\label{tab:ablations_comps}
\end{table}
\begin{table}[t]
\centering
\begin{tabular}{l|c|c}
\hline
\textbf{Model} & \textbf{EM} & \textbf{F1} \\ \hline
minus char embedding & 47.5 & 61.4 \\ 
minus question-dependent passage enc. & 32.1 & 45.0 \\ 
minus 2nd LSTM during prediction & 46.5 & 61.6 \\ 
\hline
\hline
{AMANDA} & \textbf{48.4} & \textbf{63.3} \\ \hline
\end{tabular}
\caption{Ablation of other components on the NewsQA development set}
\label{tab:other_ablations}
\end{table}
Table \ref{tab:ablations_comps} shows that AMANDA performs better than any of the ablated models which include the ablation of multi-factor attentive encoding, max-attentional question aggregation ($\mathbf{q}_{ma}$), and question type representation ($\mathbf{q}_{f}$).
We also perform statistical significance test using paired t-test and bootstrap resampling. Performance of AMANDA (both in terms of EM and F1) is significantly better $(p < 0.01)$ than the ablated models.

One of the key contributions of this paper is multi-factor attentive encoding which aggregates information from the relevant passage words by using a tensor-based attention mechanism. The use of multiple factors helps to fine-tune answer inference by synthesizing information distributed across multiple sentences. The number of factors is the granularity to which the model is allowed to refine the evidence.
The effect of multi-factor attentive encoding is illustrated by the following example taken from the NewsQA development set:

\noindent
\textit{
What will allow storage on remote servers?
}

\noindent
\textit{
...The \textbf{iCloud service} will now be integrated into the iOS 5 operating system. It will work with \underline{apps} and allow content to be stored on remote servers instead of the users' iPod, iPhone or other device...
}

\noindent
When multi-factor attentive encoding is ablated, the model could not figure out the cross-sentence co-reference and wrongly predicted the answer as {\em apps}. On the contrary, with multi-factor attentive encoding, AMANDA could correctly infer the answer as {\em iCloud service}.

Another contribution of this work is to include the question focus during prediction. It is performed by adding two components: $\mathbf{q}_{ma}$ (max-attentional question aggregation) and $\mathbf{q}_f$ (question type representation). $\mathbf{q}_{ma}$ and $\mathbf{q}_f$ implicitly infer the answer type during prediction by focusing on the important question words.
Impact of the question focus components is illustrated by the following example taken from the NewsQA development set:

\noindent
\textit{
who speaks on Holocaust remembrance day?
}
\newline
\textit{
... Israel's vice prime minister \textbf{Silvan Shalom} said Tuesday ``Israel can never ... people just 65 years ago'' ... He was speaking as \underline{Israel} observes its Holocaust memorial day, remembering the roughly...
} \newline
Without the $\mathbf{q}_{ma}$ and $\mathbf{q}_f$ components, the answer was wrongly predicted as {\em Israel}, whereas with $\mathbf{q}_{ma}$ and $\mathbf{q}_{f}$, AMANDA could correctly infer the answer type (i.e., a person's name) and predict {\em Silvan Shalom} as the answer.

Ablation studies of other components such as character embedding, question-dependent passage encoding, and the second LSTM during prediction are given in Table \ref{tab:other_ablations}. When the second LSTM ($\mathbf{E}$) is ablated, a feed-forward layer is used instead. Table \ref{tab:other_ablations} shows that question-dependent passage encoding has the highest impact on performance.
\subsection{Variation on the number of factors ($m$) and $\mathbf{q}_{ma}$}
\label{sec:variation}
\begin{table}[t]
\small
\centering
\begin{tabular}{l|c|c|c|c|c}
\hline
\textbf{Value of $m$} & \textbf{1} & \textbf{2} & \textbf{3} & \textbf{4} & \textbf{5} \\ \hline
\textbf{EM} & 45.8 & 47.4 & \textbf{48.7} & 48.4 & 48.0 \\ 
\textbf{F1} & 61.2 & 61.9 & 62.9 & \textbf{63.3} & 62.5 \\ \hline
\end{tabular}
\caption{Variation of $m$ on the NewsQA development set.}
\label{tab:variation_num_factors}
\end{table}
Table \ref{tab:variation_num_factors} shows the performance of AMANDA for different values of $m$. We use 4 factors for all the experiments as it gives the highest F1 score. Note that $m = 1$ is equivalent to standard bilinear attention. 
\begin{table}[t]
\centering
\begin{tabular}{l c c}
\hline
\textbf{Aggregation} & \textbf{EM} & \textbf{F1} \\ \hline
Mean & 46.6 & 61.3  \\ 
Sum & 47.9 & 62.2  \\ 
Max (AMANDA) & \textbf{48.4} & \textbf{63.3}  \\ \hline
\end{tabular}
\caption{Variation of question aggregation formulation on the NewsQA development set.}
\label{tab:var_q}
\end{table}

Table \ref{tab:var_q} shows the variation of question aggregation formulation. 
For mean aggregation, the attentional weight vector $\mathbf{k}$ is formulated by applying column-wise averaging on the attention matrix $\mathbf{A}$.
Intuitively, it is giving equal priority to all the passage words to determine a particular question word attention. Similarly, in the case of sum aggregation, we apply a column-wise sum operation. 
Table \ref{tab:var_q} shows that the best performance is obtained when $\mathbf{q}_{ma}$ is obtained with a column-wise maximum operation on $\mathbf{A}$. Effectively, it is helping to give higher weights to the more important question words based on the most relevant passage words.
\subsection{Quantitative Error Analysis}
\label{sec:quant_error_analysis}
\begin{figure}[t]
\subfloat[]{\includegraphics[width=3.8cm,height=2cm]{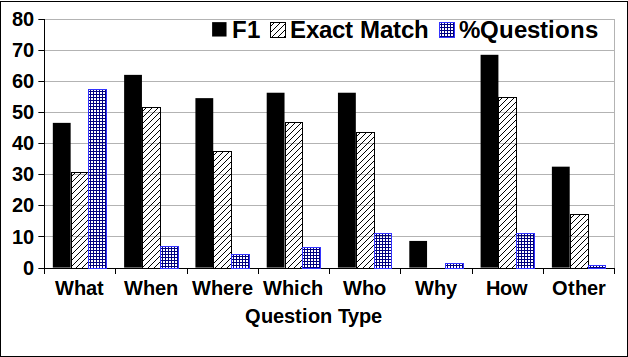}}
\hfill
\subfloat[]{\includegraphics[width=3.8cm,height=2cm]{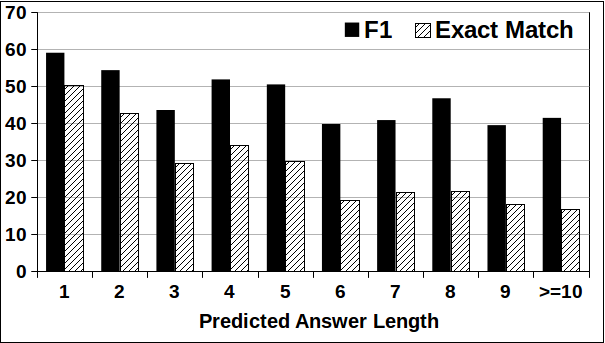}}
\caption{(a) Results for different question types. 
(b) Results for different predicted answer lengths.}
\label{fig:error_analysis}
\end{figure}
We analyzed the performance of AMANDA across different question types and different predicted answer lengths. Figure \ref{fig:error_analysis}(a) shows that it performs poorly on {\em why} and {\em other} questions whose answers are usually longer. Figure \ref{fig:error_analysis}(b) supports this fact as well. When the predicted answer length increases, both F1 and EM start to degrade. The gap between F1 and EM also increases for longer answers. This is because for longer answers, the model is not able to decide the exact boundaries (low EM score) but manages to predict some correct words which partially overlap with the reference answer (relatively higher F1 score).
\subsection{Qualitative Error Analysis}
\label{sec:qual_error_analysis}
\begin{table}[t]
\centering
\begin{tabular}{|p{8.0cm}|}
\hline
\noindent{Answer ambiguity (42\%)} \\
Q: What happens to the power supply? \\
... customers possible.'' The outages were due mostly to power \textbf{lines downed} by Saturday's hurricane-force winds, which \underline{knocked over trees and utility poles}. At  ... \\ \hline
\noindent{Context mismatch (22\%)} \\
Q: Who was Kandi Burrus's fiance? \\
Kandi Burruss, the \underline{newest cast member of the reality} \underline{show ``The Real Housewives of Atlanta}'' ... fiance, who died ... fiance, 34-year-old \textbf{Ashley ``A.J.'' Jewell}, also... \\
\hline
\noindent{Complex inference (10\%)} \\
Q: When did the Delta Queen first serve? \\
...  the Delta Queen steamboat, a floating National ... scheduled voyage \underline{this week} ... The Delta Queen will go ... Supporters of the boat, which has roamed the nation's waterways since \textbf{1927} and helped the Navy ... \\
\hline
\noindent{Paraphrasing issues (6\%)} \\
Q: What was Ralph Lauren's first job? \\
Ralph Lauren has ... Japan. For four ... than the former \textbf{tie salesman} from the Bronx. ``Those ties ... Lauren originally named his \underline{company Polo} because ... \\
\hline
\end{tabular}
\caption{Examples of different error types and their percentages. Ground truth answers are bold-faced and predicted answers are underlined.}
\label{tab:qual_analysis}
\end{table}
On the NewsQA development set, AMANDA predicted completely wrong answers on 25.1\% of the questions. We randomly picked 50 such questions for analysis. The observed types of errors are given in Table \ref{tab:qual_analysis} with examples. 42\% of the errors are due to answer ambiguities, i.e., no unique answer is present. 
22\% of the errors are due to mismatch between question and context words. 
10\% of the errors are due to the need for highly complex inference. 
6\% of the errors occur due to paraphrasing, i.e., the question is posed with different words which do not appear in the passage context.
The remaining 20\% of the errors are due to insufficient evidence, incorrect tokenization, wrong co-reference resolution, etc.
\section{Related Work}
\label{sec:related_works}
Recently, several neural network-based models have been proposed for QA.
Models based on the idea of chunking and ranking include \citeauthor{ibm_squad} (\citeyear{ibm_squad}) and \citeauthor{google_squad} (\citeyear{google_squad}). 
\citeauthor{cmu_squad} (\citeyear{cmu_squad}) used a fine-grained gating mechanism to capture the correlation between a passage and a question.
\citeauthor{smu_squad} (\citeyear{smu_squad}) used a Match-LSTM to encode the question and passage together and a boundary model determined the beginning and ending boundary of an answer. 
\citeauthor{newsqa} (\citeyear{newsqa}) reimplemented Match-LSTM for the NewsQA dataset and proposed a faster version of it.
\citeauthor{salesforce_squad} (\citeyear{salesforce_squad}) used a co-attentive encoder followed by a dynamic decoder for iteratively estimating the boundary pointers.
\citeauthor{allenai_squad} (\citeyear{allenai_squad}) proposed a bi-directional attention flow approach to capture the interactions between passages and questions.
\citeauthor{fastqa_squad} (\citeyear{fastqa_squad}) proposed a simple context matching-based neural encoder and incorporated word overlap and term frequency features to estimate the start and end pointers.
\citeauthor{rnet_squad} (\citeyear{rnet_squad}) proposed a gated self-matching approach which encodes the passage and question together using a self-matching attention mechanism.
\citeauthor{memen} (\citeyear{memen}) proposed a memory network-based multi-layer embedding model and reported results on the TriviaQA dataset.

Different from all prior approaches, our proposed multi-factor attentive encoding helps to aggregate relevant evidence by using a tensor-based multi-factor attention mechanism. This in turn helps to infer the answer by synthesizing information from multiple sentences.
AMANDA also learns to focus on the important question words to encode the aggregated question vector for predicting the answer with suitable answer type. 
\section{Conclusion}
In this paper, we have proposed a question-focused multi-factor attention network (AMANDA), which learns to aggregate meaningful evidence from multiple sentences with deeper understanding and to focus on the important words in a question for extracting an answer span from the passage with suitable answer type. AMANDA achieves state-of-the-art performance on NewsQA, TriviaQA, and SearchQA datasets, outperforming all prior published models by significant margins. Ablation results show the importance of the proposed components.
\section{Acknowledgement}
This research was supported by research grant R-252-000-634-592.
\bibliographystyle{aaai} 
\bibliography{aaai2018}
\end{document}